\documentclass[conference]{IEEEtran}
\IEEEoverridecommandlockouts

\usepackage{amsmath}
\usepackage{amssymb}
\usepackage{amsfonts}
\usepackage{graphicx}
\usepackage{textcomp}
\usepackage{xcolor}
\usepackage{tcolorbox}
\tcbuselibrary{breakable}
\usepackage{fvextra}
\usepackage{url}
\usepackage{needspace}
\usepackage{xspace}
\usepackage{cite}
\usepackage{stfloats}
\usepackage{pifont}
\newtcolorbox{promptbox}[1][]{
  colback=blue!3!white, colframe=blue!45!black,
  coltitle=white, colbacktitle=blue!45!black,
  boxrule=0.5pt, arc=2pt, left=2pt, right=2pt, top=2pt, bottom=2pt,
  before skip=6pt, after skip=6pt, breakable,
  fonttitle=\sffamily\bfseries\small, #1
}

\makeatletter
\def\abstract{\normalfont\@IEEEtweakunitybaselinestretch{1.15}
    \if@twocolumn
      \@IEEEabskeysecsize\noindent\textit{\abstractname}---\relax
    \else
      \bgroup\par\addvspace{0.5\baselineskip}\centering\vspace{-1.78ex}\@IEEEabskeysecsize\textit{\abstractname}\par\addvspace{0.5\baselineskip}\egroup\quotation\@IEEEabskeysecsize
    \fi\@IEEEgobbleleadPARNLSP}
\makeatother

\usepackage[hidelinks]{hyperref}

\hypersetup{
pdftitle={FedSkillGit: Federated Skill Evolution via Version Control},
pdfauthor={Boxun Lan, Jiamei Li, Weijian Li, Chengyu Yang, Zhuang Qi, Chengcheng Wang, Xiaohong Chen, Ming Hu},
pdfsubject={Federated learning, skill evolution, large language model agents},
pdfkeywords={federated learning, skill evolution, large language model agents},
}

\usepackage{amsmath,amsfonts,bm}

\def\eqref#1{equation~\ref{#1}}

\def\1{\bm{1}}

\DeclareMathAlphabet{\mathsfit}{\encodingdefault}{\sfdefault}{m}{sl}
\SetMathAlphabet{\mathsfit}{bold}{\encodingdefault}{\sfdefault}{bx}{n}

\newcommand{\vunion}{V_{\cup}}

\newcommand{\ourmethod}{FedSkillGit\xspace}
\newcommand{\ourmethodbf}{\textbf{\ourmethod}}

\title{FedSkillGit: Federated Skill Evolution via Version Control}

\author{
\IEEEauthorblockN{Boxun Lan\textsuperscript{1}, Jiamei Li\textsuperscript{1},
Weijian Li\textsuperscript{1}, Chengyu Yang\textsuperscript{1},
Zhuang Qi\textsuperscript{2},\\ Chengcheng Wang\textsuperscript{1},
Xiaohong Chen\textsuperscript{1}, Ming Hu\textsuperscript{1*}}
\IEEEauthorblockA{\textsuperscript{1}East China Normal University, Shanghai, China\\
\textsuperscript{2}Shandong University, Jinan, China\\
\textsuperscript{*}Corresponding author: Ming Hu, mhu@sei.ecnu.edu.cn}
}

\begin{document}

\maketitle

\begin{abstract}

Large language model agents can improve their capabilities by evolving reusable natural-language skills from execution experience. In real-world deployments this experience is distributed across data owners who cannot pool their raw data, which motivates federated skill evolution. The natural first design lets each client evolve the whole skill document and upload the result for server-side aggregation. That design is poorly matched to skill evolution for three reasons. An evolved document can still carry client-specific content distilled from private traces, repeated transmission of a nearly unchanged document spends communication on knowledge that the population already holds, and the contribution of a single client is buried in a document that the server cannot meaningfully attribute. We propose \ourmethodbf, a federated skill evolution framework that replaces full-document updates with commits. Each participating client submits a bounded set of patches over named rules together with its local experience, which summarizes what the round taught it, and the pair passes a two-stage privacy filter before it leaves the client. The server deduplicates the submitted patches and the local experience of all participants, updates a global experience document that only the optimizers read, and merges the surviving patches into a candidate skill. The candidate is distributed to the clients, which score it on their own validation sets and report the results, and it is published only when the reported scores improve on the current version. Experiments on three datasets and three language model families show that \ourmethod outperforms state-of-the-art baselines by up to 22\% and approaches centralized evolution without sharing raw client data.

\end{abstract}


\section{Introduction}

Large language model (LLM) agents can improve their task performance by reusing knowledge acquired from previous executions rather than solving similar tasks repeatedly. 
To support such knowledge reuse, recent approaches distill successful trajectories, failure feedback, and task-specific experience into natural-language skills that guide subsequent reasoning and execution \cite{AutoSkill, SkillPyramid, SkillOpt}. 
As agents encounter new tasks and environments, these skills need to be continuously refined to incorporate the new acquired knowledge. 
Therefore, effective skill evolution is important for agents that are expected to continuously improve through experience.

Existing skill evolution methods primarily rely on the execution experience accumulated by individual agents, limiting skill improvement to the scope of local experience.
In real-world deployments, agents operating across diverse tasks and execution environments can collectively accumulate a broader range of experiences, providing richer knowledge to support skill evolution.
However, directly sharing the execution trajectories that generated these experiences risks exposing sensitive input data, private environmental state, or proprietary workflows.
Therefore, how to achieve collaborative skill evolution using distributed experiences without sharing raw execution trajectories has become an important research issue.

Federated learning \cite{FedAvg, Kairouz, fedmut, hu2023gitfl, fedfsa, lu2025gradients, huang2024federated} provides a natural paradigm for such collaboration by enabling multiple clients to jointly improve a shared model without exchanging their raw data. 
Following this paradigm, a straightforward approach to federated skill evolution is to allow each client to refine a shared skill using its local execution experience and upload the evolved skill document to the server for aggregation. 
However, unlike conventional FL, which aggregates numerical model updates, skill evolution involves structured natural-language knowledge. Directly adopting the traditional FL paradigm faces three key challenges as follows.
\ding{182} First, keeping raw traces local does not necessarily preserve privacy because an evolved skill can still encode client-specific operations, entities, or environmental information distilled from those traces. 
\ding{183} Second, local evolution often modifies only a small fraction of the shared skill, while repeatedly transmitting the entire document causes communication to grow with the accumulated skill rather than with the newly acquired knowledge.
\ding{184} Third, the actual knowledge contributed by a client is buried in a largely unchanged document, making it difficult for the server to identify, evaluate, and combine useful revisions across clients. 
These limitations reveal a mismatch between the update abstraction of conventional federated learning and the structured natural-language artifacts optimized by skill evolution.
Therefore, \textit{how to represent and aggregate client knowledge in a privacy-preserving, communication-efficient, and fine-grained manner is an important challenge for federated skill evolution.}

To address these challenges, inspired by the incremental revision mechanism of Git in collaborative software development, we propose \ourmethod, a novel federated skill evolution paradigm. Instead of aggregating full skill documents, clients exchange semantic commits containing fine-grained skill revisions and supporting experience, enabling efficient knowledge exchange and integration. To support this process, \ourmethod incorporates two-stage privacy filtering to mitigate potential information leakage, a global experience document to guide local skill evolution and server-side aggregation, and version validation to control the publication of updated skills. The main contributions of this paper are summarized as follows.
% \begin{itemize}
% \item We propose a Git-like federated skill evolution framework, named \ourmethodbf, which adopts a commit-driven skill evolution mechanism to facilitate the collaborative skill optimization among multiple clients.
% \item We introduce a cross-round experience reuse mechanism. It avoids redundant exploration of revisions the population has already rejected.
% \item We introduce a sanitize-then-audit privacy filter that operates on the commit, which addresses the privacy risk of transmitting natural-language edits.
% \item Extensive experiments on three datasets and three language model families demonstrate that \ourmethod outperforms the strongest baseline by up to 22\% while approaching centralized performance.
% \end{itemize}

\begin{itemize}
    \item We propose \ourmethodbf, a novel federated skill evolution framework that integrates fine-grained commit aggregation and version validation to enable reliable collaborative evolution of a shared skill.
    \item We design a cross-round experience reuse mechanism that leverages accumulated global experience to guide local skill optimization and server-side aggregation, reducing redundant knowledge exploration and integration.
    \item We develop a two-stage privacy filtering mechanism that combines sanitization and private-value auditing to mitigate the risk of sensitive information leakage.
    \item Extensive experiments on three datasets and three language model families demonstrate that \ourmethod outperforms the strongest baseline by up to 22\% while approaching centralized performance.
\end{itemize}
\section{Related Work}

\subsection{Federated learning}

Federated learning (FL) \cite{FedAvg} enables multiple clients to collaboratively train global model while keeping their raw data local. 
Existing FL studies have extensively investigated data and system heterogeneity in distributed scenarios.
To address data heterogeneity, existing studies have explored various approaches for learning a shared global model, including proximal regularization \cite{FedProx}, control variates \cite{SCAFFOLD}, knowledge distillation \cite{FedDF}, prototype-based aggregation \cite{cao2026two,fu2025Beyond}, and adaptive aggregation \cite{fedmr,fedcross}. 
Personalized FL methods instead optimize client-specific models to accommodate heterogeneous local data distributions \cite{PerFedAvg,pFedMe}. 
To address system heterogeneity, other studies investigate heterogeneous model architectures and resource-constrained client participation \cite{FedCS,adaptivefl,liu2025fedgraft,xia2025multisfl}. 
Despite these advances, FL methods primarily exchange model parameters or other model-related representations, rather than structured natural-language skills.

\subsection{Skill evolution}
LLM agents increasingly employ skill libraries or documents to preserve and reuse knowledge acquired from prior experience. Accordingly, recent studies have investigated various approaches to automated skill acquisition, organization, and evolution. 
For skill acquisition, existing methods extract reusable knowledge from execution trajectories \cite{AutoSkill, EmbodiSkill, Memento-Skills, XSkill}. 
To improve knowledge organization and reuse, other studies explore hierarchical skill organization \cite{SkillPyramid, SkillNet}, redundancy reduction through compression and abstraction \cite{SkillReducer, SkillBank}, and large-scale skill curation and composition \cite{SkillX, SkillOS, SkillFoundry, SkillForge, Skills-Coach}. 
Beyond skill acquisition and management, recent research further investigates skill refinement and evolution through co-evolution and verification \cite{CoEvoSkills}, parallel generation and consolidation \cite{Trace2Skill}, reinforcement learning \cite{Skill1, SkillRL, SPARK}, and agentic evolution processes \cite{SkillClaw}. 
Complementary research also examines the reusable capabilities that language models acquire from in-context experience \cite{Context-to-Skills}.

However, most existing skill evolution methods assume that the experience used for evolution is available to a single owner or can be centrally accessed. As a result, they do not address how knowledge distributed across privacy-constrained data owners can be collaboratively incorporated into a shared skill. Moreover, unlike conventional federated learning where client updates are typically represented as numerical parameters or gradients, skill evolution directly modifies structured natural-language artefacts such as rules, procedures, and local patches. This distinction motivates aggregation mechanisms that operate over discrete revisions rather than numerical updates.

\subsection{Federated skill evolution}
Recent work has begun to extend federated collaboration from model parameters to agent evolution. In particular, FederatedSkill \cite{FederatedSkill} enables agents to collaboratively evolve skill libraries without sharing raw execution trajectories. It represents local experience as skill diffs containing structured modifications to local skill libraries and aggregates these diffs through a server-side evolution agent. Its primary objective is personalized skill evolution under client heterogeneity, where different clients may use different backbone models, agent harnesses, and task distributions. Accordingly, the server infers client-specific capability boundaries from uploaded patches and constructs a personalized skill library for each client.

Unlike FederatedSkill, which maintains personalized skill libraries for heterogeneous clients with heterogeneous tasks, \ourmethod aims to continuously evolve a generalized skill from distributed private experience for similar tasks. 
This setting requires determining which locally discovered revisions should become part of the shared knowledge and whether their composition improves the resulting skill. \ourmethod therefore treats collaborative skill evolution as a version-controlled knowledge integration process, where commits are individually evaluated and selectively merged into a shared version, and each new version is validated and rolled back when aggregation causes regression. Our focus is thus not client-specific personalization, but reliable accumulation and validation of distributed knowledge within a common evolving skill.

\section{Method}
\label{sec:method}

In this section, we introduce the proposed \ourmethod framework. We first provide an overview of the framework, followed by the local skill evolution, which includes commit generation and privacy filtering. Finally, we describe the server-side skill updating process, consisting of experience updating, experience-aware aggregation, and version validation.

\begin{figure*}[b]
  \centering
  \includegraphics[width=\linewidth]{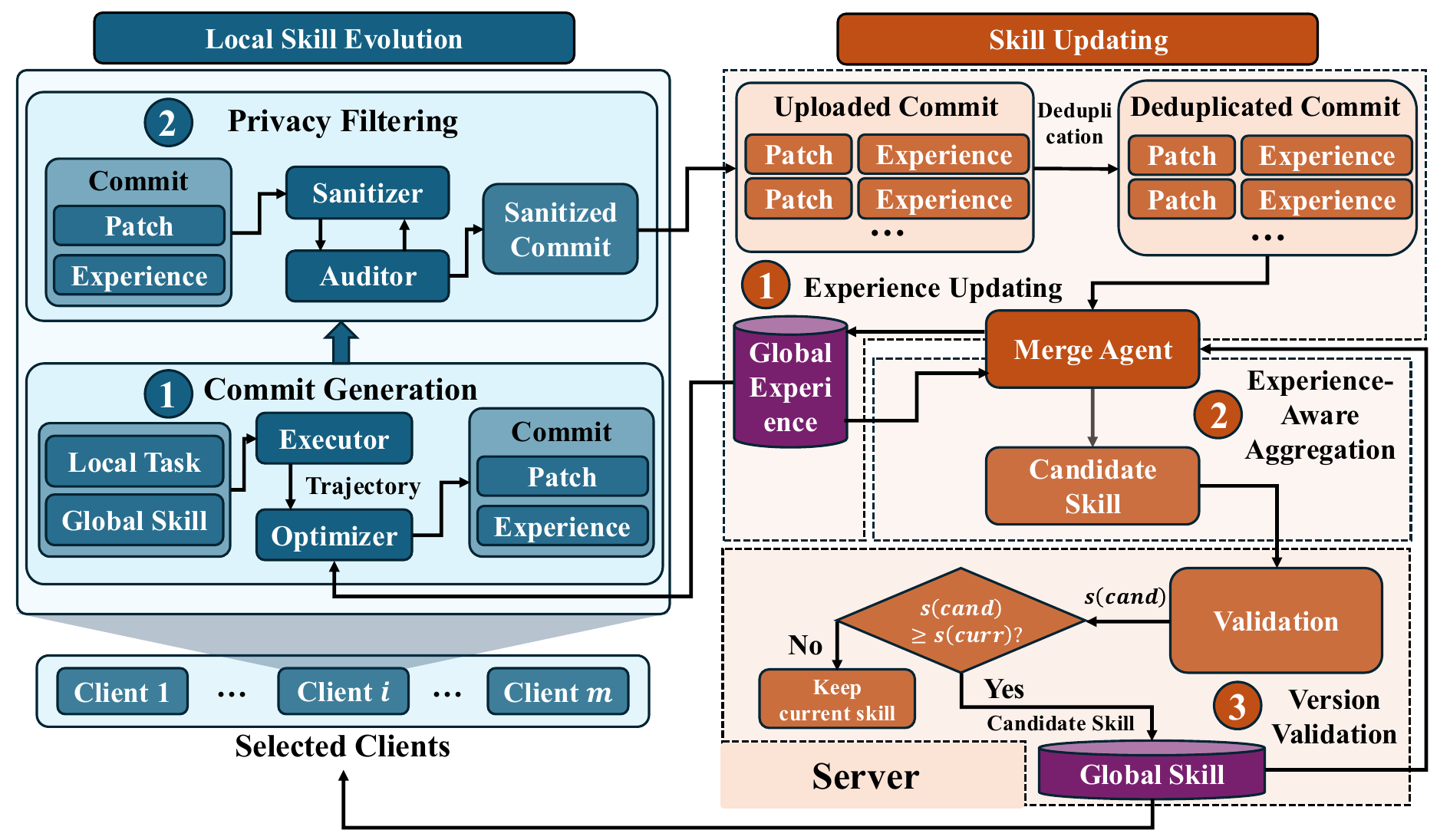}
  % \caption{Overview of our proposed \ourmethod. The server distributes the shared skill $d$ to
  % the selected clients. A client runs the executor on its local tasks to collect
  % trajectories, and the optimizer agent reads these trajectories together with
  % the global experience document $e$ to produce a patch and a local experience.
  % Both are packaged as a commit, which passes the privacy filter only after the
  % sanitizer agent rewrites it and the audit agent confirms that it carries no
  % value from the client's private-value set. The server deduplicates the uploaded
  % commits, uses the reported experience to update $e$, and merges the patches into
  % a candidate skill. The candidate is validated on the union of the clients'
  % validation sets and replaces $d$ only if it improves the score.}
  \caption{Overview of our proposed \ourmethod. The framework consists of local skill evolution and server-side skill updating. Each selected client generates a commit containing skill patches and local experience from execution trajectories, followed by two-stage privacy filtering. The server deduplicates uploaded commits, updates global experience, and aggregates patches into a candidate skill. Finally, version validation determines whether to publish the candidate based on its performance.}
  \label{fig:method}
\end{figure*}

\subsection{Federated framework}
\label{sec:sys}
We consider a federated skill evolution system consisting of a central server and $N$ clients, denoted by $\mathcal{C}=\{1,2,\ldots,N\}$.
The server maintains a shared natural-language skill document $d$ and a global experience document $e$, which jointly support collaborative skill evolution.
Each client $i\in\mathcal{C}$ maintains a private training dataset $\mathcal{D}_i$ and a local validation dataset $\mathcal{V}_i$, with heterogeneous task distributions across clients. 
The training dataset is used for local task execution and skill refinement, while the validation dataset is reserved for evaluating candidate skills. 
Each client is equipped with four LLM-based agents, including an executor $\mathbf{E}$, a skill optimizer $\mathbf{O}$, a sanitizer $\mathbf{S}$, and an auditor $\mathbf{A}$, where $\mathbf{E}$ and $\mathbf{O}$ are responsible for local task execution and skill refinement, while $\mathbf{S}$ and $\mathbf{A}$ perform privacy filtering on the generated commits. 
On the server, a merge agent $\mathbf{M}$ is employed to update the global experience document and aggregate the uploaded skill patches, while version validation determines whether the resulting candidate skill is accepted as the new global skill version.

The federated evolution proceeds over multiple communication rounds. 
Specifically, in each communication round $r$, a set of $m$ participating clients denoted by $\mathcal{P}_r \subseteq \mathcal{C}$ is selected for skill evolution.
Participating clients generate skill updates from their private execution experience and transmit privacy-filtered commits to the server without sharing raw task data or execution trajectories. The server integrates these skill updates to produce a candidate skill, which is evaluated using the clients' local validation datasets for version control. The objective is to collaboratively improve the task performance of a single shared global skill across heterogeneous clients while keeping their private execution data local.

\subsection{Overview}
Fig.~\ref{fig:method} illustrates the overall framework of our proposed \ourmethod, which consists of local skill evolution on the clients and skill updating on the server. In each communication round, the selected clients refine the shared global skill using their private execution experience and submit semantic commits to the server. The server then integrates these commits to update the global experience and generate candidate skill, which undergoes version validation and determines whether to update the global skill. Through this iterative process, \ourmethod enables collaborative skill evolution across distributed clients without sharing raw execution data.

\textbf{Local Skill Evolution.} In each communication round, the server distributes the current global skill 
% and global experience document
to the selected clients. Each client employs an executor to perform local tasks under the guidance of the current global skill and generate execution trajectories.
Based on these trajectories and the global experience, the optimizer produces a rule-level skill patch and a local experience, which are organized into a semantic commit.
Since the generated commit may still contain sensitive information derived from private execution data, it proceeds through a two-stage privacy protection process consisting of sanitization and auditing. The sanitizer removes potentially sensitive client-specific information, and the audit agent examines the sanitized commit. Only commits that pass the privacy audit are transmitted to the server.

\textbf{Skill Updating.} Upon receiving the client commits, the server performs three stages of skill updating, including experience updating, experience-aware aggregation, and version validation. First, the server deduplicates the uploaded patches and experiences and incorporates the newly acquired knowledge into the global experience document. The merge agent then integrates the remaining patches into the current global skill, guided by the accumulated experience of previous rounds, to generate a candidate skill. The global experience document also informs subsequent local optimization by recording previously observed failure patterns and attempted revisions, reducing redundant exploration and aggregation. Notably, this document is accessed only by the optimizer and merge agents, while the executor relies exclusively on the shared skill for task execution. Finally, the candidate skill is evaluated by clients on their private validation datasets, and only the resulting success rates are returned to the server. The candidate is published as the new global skill only when its aggregate validation performance is no lower than that of the current version; otherwise, the current skill is retained. The evolution process terminates after two consecutive rounds without an accepted update.

\subsection{Local Skill Evolution}
\label{sec:local}
The local skill evolution process consists of commit generation and privacy filtering two stages. 
In the commit generation stage, each participating client extracts skill patch and experience summaries from its private execution experience to construct a commit. 
In the privacy filtering stage, the generated commit undergoes privacy filtering before transmission to mitigate potential information leakage.
\subsubsection{Commit generation}
In communication round $r$, each participating client $i \in \mathcal{P}_r$ receives the current global skill $d_{r-1}$ and executes its local training set $\mathcal{D}_i$ using its own executor.
Based on the current global skill $d_{r-1}$, the executor generates task responses and their corresponding execution trajectories. 
Subsequently, the skill optimizer $\mathbf{O}$ analyzes the trajectories with the global experience document $e_{r-1}$ to extract its local patch and experience.
Specifically, the local skill patch can be represented as a sequence of rule-level edits
\begin{equation}
    p_i^r=(\delta_{i,1}^r,\ldots,\delta_{i,j}^r,\ldots,\delta_{i,K_i^r}^r),
\end{equation}
where $K_i^r$ denotes the number of edits proposed by client $i$ in round $r$.
Each edit specifies a modification to a particular rule in the current skill document, formally defined as
\begin{equation}
\delta_{i,j}^r = (o_{i,j}^r, t_{i,j}^r, u_{i,j}^r),
\end{equation}
where $o_{i,j}^r$, $t_{i,j}^r$, and $u_{i,j}^r$ denote the editing operation, target rule, and revision content, respectively.
To constrain the scope of local modifications, the editing operations include \texttt{append}, \texttt{insert-after}, and \texttt{replace}, which can be formulated as
\begin{equation}
o_{i,j}^r \in
\{\texttt{append},\texttt{insert-after},\texttt{replace}\}.
\end{equation}
Meanwhile, the local experience summary $e_i^r$ captures observed failure patterns and the execution experience motivating the proposed revisions.
Finally, the generated skill patch and local experience are organized into a semantic commit, which is represented as
\begin{equation}
c_i^r = (p_i^r, e_i^r).
\end{equation}

\subsubsection{Privacy filtering}
\label{sec:privacy}
Although raw execution trajectories remain local, the generated commits may contain sensitive information derived from private training data. 
To mitigate privacy risk, \ourmethod designs a two-stage privacy filtering mechanism consisting of sanitization and auditing. 

In the first stage, an LLM-based sanitizer $\mathbf{S}$ transforms each generated commit according to predefined sanitization instructions, formally expressed as
\begin{equation}
\tilde{c}_i^r =
\mathcal{S}(c_i^r;\mathcal{I}_{\mathrm{san}})
= (\tilde{p}_i^r,\tilde{e}_i^r),
\end{equation}
where $\mathcal{S}$ denotes the sanitization operation guided by instructions $\mathcal{I}_{\mathrm{san}}$, and $\tilde{p}_i^r$ and $\tilde{e}_i^r$ represent the sanitized skill patch and local experience, respectively.
Specifically, the sanitizer identifies potentially sensitive information in the free-text fields, including client-specific entities, identifiers, and environment details, and replaces them with typed placeholders.
To preserve the applicability of skill revisions, each sanitized edit retains its original operation and target rule, expressed as
\begin{equation}
\tilde{\delta}_{i,j}^r =
(o_{i,j}^r,t_{i,j}^r,\tilde{u}_{i,j}^r),
\end{equation}
where $\tilde{u}_{i,j}^r$ denotes the sanitized revision content.

In the second stage, the sanitized commit undergoes a privacy audit against a client-specific private-value set $\mathcal{Q}_i$, which is extracted from local training data and remains fixed throughout the evolution process.
The auditor $\mathbf{A}$ examines the sanitized commit to detect potential disclosures, formally expressed as
\begin{equation}
\mathbf{A}(\tilde{c}_i^r, \mathcal{Q}_i) =
\begin{cases}
1, & \text{if } \forall v \in \mathcal{Q}_i,\,
\neg \operatorname{Match}(v,\tilde{c}_i^r),\\
0, & \text{otherwise},
\end{cases}
\end{equation}
where $\operatorname{Match}(v,\tilde{c}_i^r)$ indicates whether a private value $v$ appears in the free-text fields of the sanitized commit.
If the audit fails, the commit is returned to the sanitization stage for further revision.
Only commits that pass the privacy audit are transmitted to the server.

The two stages serve complementary purposes. 
Sanitization removes potentially sensitive content while preserving the structure of skill revisions, whereas auditing provides an additional verification step based on a fixed set of client-specific private values. 
This mechanism reduces the risk of disclosing sensitive information through transmitted commits. 

\subsection{Skill Updating}
\label{sec:update}
After the server receives sensitivity-protected commits uploaded by clients, the next global skill is determined through experience updates, experience-based aggregation, and version verification three steps.

\subsubsection{Experience Updating}
The server first extracts the local experience summaries from the uploaded commits $\{c_i^r \mid i \in \mathcal{P}_r\}$ and performs deduplication to eliminate redundant or conflicting information, which can be expressed as
\begin{equation}
\bar{c}_r =
\operatorname{Dedup}
\left(\{c_i^r \mid i \in \mathcal{P}_r\}\right),
\end{equation}
where $\bar{c}_r=(\bar{p}_r, \bar{e}_r)$ represents the deduplicated local commits.
Subsequently, the merge agent $\mathbf{M}$ integrates the experiences in these commits with the current global experience document $e_{r-1}$ to obtain the updated global experience, that is
\begin{equation}
e_r = \mathbf{M}_{\mathrm{exp}}
\left(e_{r-1}, \bar{e}_r\right).
\end{equation}
The accumulated experience is organized into typed entries to facilitate structured knowledge management and subsequent retrieval.

\subsubsection{Experience-aware aggregation}
After updating the global experience document, the server performs experience-aware aggregation to construct a candidate skill.
Specifically, the merge agent $\mathbf{M}$ then integrates the deduplicated patches based on the current global skill $d_{r-1}$ under the guidance of the updated global experience document $e_r$. Here, the candidate skill $\tilde{d}_r$ can be formulated as
\begin{equation}
\tilde{d}_r =
\mathbf{M}(d_{r-1}, e_r).
\end{equation}

During aggregation, the merge agent examines the submitted edits and classifies them as kept, deduplicated, promoted to shared rules, or rejected as invalid.
The global experience document provides accumulated knowledge about previously observed failure patterns and attempted revisions, enabling the merge agent to identify redundant or previously rejected proposals.
Consequently, redundant or invalid edits are excluded, while the remaining revisions are incorporated into the current global skill.
Unlike conventional federated learning that aggregates numerical model updates, this process integrates natural-language revisions while preserving their intended modifications and allowing incorporated changes to be traced to their original proposals.

\subsubsection{Version validation}
\label{sec:validation}
After experience-aware aggregation, the candidate skill $\tilde{d}_r$ undergoes version validation before publication. 
Since clients operate on heterogeneous task distributions, the skill revisions that improve performance on individual clients may not necessarily benefit the entire clients. 
Therefore, a population-level validation mechanism is employed to determine whether the candidate skill $\tilde{d}_r$ can be the next global skill version.
Specifically, the server distributes the candidate skill $\tilde{d}_r$ to all clients, which independently evaluate its task success rate on their own local validation set and then return the corresponding evaluation results.
The server aggregates these results to obtain the global evaluation score $s(\tilde{d}_r)$, which is compared with the previously recorded score $s(d_{r-1})$.
The candidate skill $\tilde{d}_r$ is published only when its evaluation score is no lower than that of the current global skill $d_{r-1}$. 
Otherwise, the candidate skill is rejected, and the previous global skill is retained. 
In this case, the next global skill $d_r$ can be formulated as
\begin{equation}
d_r =
    \begin{cases}
    \tilde{d}_r, & s(\tilde{d}_r) \geq s(d_{r-1}),\\
    d_{r-1}, & \text{otherwise}.
    \end{cases}
\end{equation}
Through this validation mechanism, the published global skill maintains a non-decreasing evaluation score across communication rounds, while clients retain their private validation data. 
In addition, the evolution process terminates when candidate skills are rejected in two consecutive communication rounds. 
To characterize the progression of skill evolution, the convergence round $r^c$ is defined as the last round in which a candidate skill is accepted. 
\section{Experiments}
\label{sec:exp}
In this section, we first introduce the experimental setup and then evaluate the performance of the proposed method against baselines. 
Subsequently, ablation studies are conducted to investigate the effectiveness of individual components.
Finally, we examine the scalability of the proposed method under different numbers of clients and participation ratios.
\subsection{Experimental setup}
We consider a federated skill evolution system consisting of 10 clients with a participation ratio of 30\%, where three clients are selected in each communication round. Experiments are conducted using GPT-5.4-mini, Qwen3.5-flash, and Gemini3-flash large language models on three datasets, including SearchQA for open-domain question answering, LiveMath for research-level multiple-choice mathematical problems, and DocVQA for document-based visual question answering. Each dataset provides an initial seed skill document $d_0$.

Task success rate (TSR) is adopted as the primary evaluation metric and measured on held-out test sets using the evaluator provided by each dataset. We compare \ourmethod with four baselines, including Centralized, No Skill, Independent, and DocMerge. For a fair comparison, consistent per-round budgets and publication criteria are applied across applicable methods, while the same executor and merging mechanisms are employed where relevant.

\begin{itemize}
    \item \textit{Centralized.} One process is trained on the pooled per-round exposure, which is the same per-round data that the federated method receives, and its candidate versions are evaluated on the union of validation sets.
    \item \textit{No skill.} The frozen executor is queried with no skill document, which separates what the model already knows from what the evolved document adds.
    \item \textit{Independent.} Each client evolves its own copy of the seed on its own data with no collaboration, and the reported value is the mean over clients.
    \item \textit{DocMerge.} Each client uploads exactly one mature document and a strong optimizer merges the set in a single call, which is then evaluated on the clients' validation union. This baseline isolates iterative scheduling against a single-shot merge.
\end{itemize}

\subsection{Performance Comparison}

\begin{table}[!htbp]
  \centering
  \caption{The TSR results comparison for five methods applied to three datasets and three models with 10 clients at 30\% participation rate.(\%)}
  \label{tab:main}
  \begin{tabular}{ccccc}
    \noalign{\hrule height 1pt}
    Model & Method & SearchQA & LiveMath & DocVQA \\
    \hline
    & Centralized & 79.4 & 28.8 & 86.4 \\
    \cline{2-5}
    & No skill & 75.9 & 14.7 & 70.9 \\
    GPT-5.4-mini & Independent & 76.1 & 27.2 & 82.8 \\
    & DocMerge & 77.5 & 28.0 & 81.3 \\
    & \ourmethodbf & \textbf{78.6} & \textbf{28.8} & \textbf{88.5} \\
    \hline
    & Centralized & 83.1 & 32.8 & 90.4 \\
    \cline{2-5}
    & No skill & 77.7 & 23.2 & 77.6 \\
    Qwen3.5-flash & Independent & 80.1 & 25.6 & 83.7 \\
    & DocMerge & 81.2 & 27.2 & 80.2 \\
    & \ourmethodbf & \textbf{82.4} & \textbf{29.6} & \textbf{85.3} \\
    \hline
    & Centralized & 85.9 & 41.6 & 91.4 \\
    \cline{2-5}
    & No skill & 79.2 & 31.2 & 86.6 \\
    Gemini3-flash & Independent & 77.2 & 31.2 & 88.4 \\
    & DocMerge & 83.1 & 32.8 & 88.4 \\
    & \ourmethodbf & \textbf{87.3} & \textbf{40.0} & \textbf{90.1} \\
    \noalign{\hrule height 1pt}
  \end{tabular}
\end{table}
Table~\ref{tab:main} presents the TSR results of different methods across three datasets and three models.

The no-skill row gives the reference for the gain that an evolved document provides. 
Federating the shared document outperforms independent evolution, since \ourmethod exceeds Independent in all nine cells by between +1.6, on GPT-5.4-mini and LiveMath, and +10.1, on Gemini3-flash and SearchQA, with a mean gain of +4.3. 
\ourmethod also recovers most of the benefit of centralizing, with a median gap of $-0.8$ against Centralized and a lead in two of the nine cells. By family average it leads on GPT-5.4-mini by +0.4 and trails on Gemini3-flash by $-0.5$ and on Qwen3.5-flash by $-3.0$. Single-shot merging is not sufficient, since DocMerge trails in all nine cells by between +0.8 and +7.2. One aggregation call has to reconcile every submitted edit at once and no later round can correct a poor reconciliation.

The margins vary with how well a single client's slice represents the task. On LiveMath, where the instances are narrow in scope, \ourmethod leads Independent by 1.6 and 4.0 points for GPT-5.4-mini and Qwen3.5-flash. On DocVQA, where instances are the most heterogeneous across clients, it leads all three model families. The gap to the Centralized reference remains largest on Gemini3-flash.

\subsection{Ablation Studies}

\subsubsection{Effectiveness of the global experience document}

\begin{table}[!htbp]
  \centering
  \caption{Ablation of the global experience document on GPT-5.4-mini and SearchQA. Task success rate (\%).}
  \label{tab:experience}
  \begin{tabular}{lc}
    \noalign{\hrule height 1pt}
    Variant & TSR (\%) \\
    \hline
    Full FedSkillGit & 78.6 \\
    w/o experience in next-round training & 76.9 \\
    w/o experience in the federated merge & 77.2 \\
    \noalign{\hrule height 1pt}
  \end{tabular}
\end{table}

Table~\ref{tab:experience} isolates the global experience document on GPT-5.4-mini and SearchQA. Both ablated variants keep the same commits, merge operator and publication gate, and they differ only in whether the accumulated experience reaches the optimizer in the next round and reaches the merge agent. Removing either use lowers task success rate relative to the full protocol, from 78.6 to 76.9 and 77.2, so the global experience document contributes in both roles. 

At merge time the global experience document tells the merge agent which submitted edits the population has already tried and rejected, so a proposal that looks attractive in isolation can be recognized as a repetition. At training time they tell the optimizer which failures the previous document failed to address on that client's slice, so the next proposal starts from the residual error instead of rediscovering it. 

\subsubsection{Effectiveness of the sanitizer}

\begin{table}[!htbp]
  \centering
  \caption{Privacy ablation (GPT-5.4-mini + SearchQA). Share of payloads that pass the audit on the first attempt, with and without the sanitizer.}
  \label{tab:privacy}
  \begin{tabular}{lc}
    \noalign{\hrule height 1pt}
    Setting & Audit first-pass rate (\%) \\
    \hline
    With sanitizer & 81 \\
    Without sanitizer & 33 \\
    \noalign{\hrule height 1pt}
  \end{tabular}
\end{table}

Table~\ref{tab:privacy} evaluates the two-stage privacy filter of Section~\ref{sec:method} on the payloads produced by real runs, again on GPT-5.4-mini and SearchQA. Both rows report first-pass rates, so the table isolates the effect of sanitization. With sanitization in place, 81 \% of payloads clear the audit on the first attempt, against 33 \% when it is removed, so the share of rejected payloads falls from 67 \% to 19 \%. 

\begin{figure}[!ht]
  \centering
  \begin{minipage}[t]{0.45\linewidth}
    \centering
    \includegraphics[width=\linewidth]{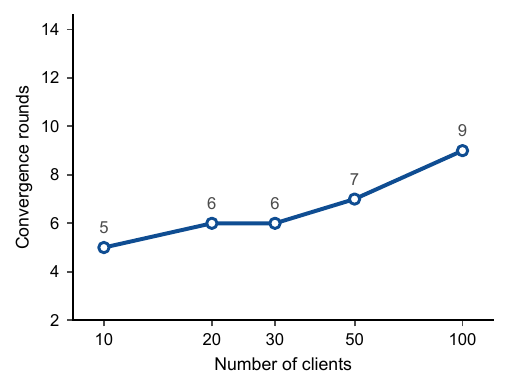}
    \caption{Convergence round count against the number of clients, with the
    participation ratio fixed at 30\,\% on GPT-5.4-mini and SearchQA. The count
    rises monotonically and the increase accelerates with population size.}
    \label{fig:rounds-clients}
  \end{minipage}\hfill
  \begin{minipage}[t]{0.45\linewidth}
    \centering
    \includegraphics[width=\linewidth]{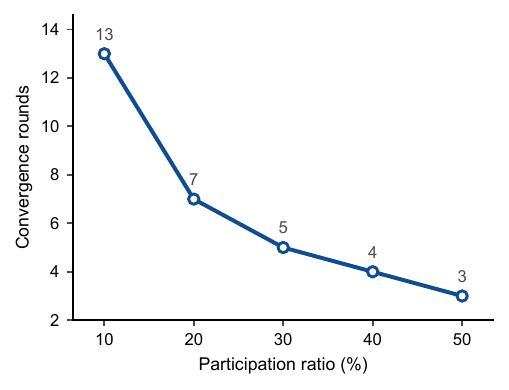}
    \caption{Convergence round count against the participation ratio, with ten
    clients on GPT-5.4-mini and SearchQA. Denser participation shortens the path,
    with strongly diminishing returns.}
    \label{fig:rounds-participation}
  \end{minipage}
\end{figure}

\subsection{Scalability}

\subsubsection{Impact of the number of clients}

Fig.~\ref{fig:rounds-clients} plots the convergence round count against the number of clients. The count rises monotonically from five rounds at ten clients to nine rounds at one hundred clients, and the increase accelerates with population size. Raising the population from ten to thirty clients adds one round, from 5 to 6, whereas raising it from fifty to one hundred adds two, from 7 to 9. The pattern follows from how much of the document remains unresolved. In the early rounds, most submitted patches are accepted, so each round yields a large improvement. As the document matures, the remaining failures are those on which clients disagree, so submissions increasingly overlap, fewer patches survive the merge, and more rounds are needed before the two consecutive rejections that terminate training are observed.

\subsubsection{Impact of the participation ratio}

Fig.~\ref{fig:rounds-participation} reports the complementary measurement. Raising the participation ratio from 10 \% to 50 \% shortens the path from 13 rounds to 3, because a larger participating subset contributes more evidence in each round. The returns diminish sharply. Raising the ratio from 10 \% to 20 \% removes six rounds, from 13 to 7, whereas raising it from 40 \% to 50 \% removes one, from 4 to 3.

\section{Conclusion}

We presented FedSkillGit, which evolves a single shared skill document  of a population of clients with experience transference and privacy protection.
Clients exchange bounded commits rather than complete documents, each commit is
filtered before transmission, and the server aggregates the surviving patches
into one evolving skill, guided by a global experience document that records the
failure patterns and rejected revisions of earlier rounds.

Across three language model families and three datasets, federating the shared
document improves task success over independent evolution in every cell by
between 1.6 and 10.1 points, with a mean gain of 4.3 points, and it lands within
about one point of centralized evolution, with a median gap of $-0.8$. No raw
execution data leaves a client at any point in the protocol. 
The convergence cost is predictable enough to
inform deployment choices, since it rises with the number of clients, from five
rounds at ten clients to nine at one hundred, and falls with the participation
ratio, from thirteen rounds at 10 \% to three at 50 \%.

\textit{Outlook.} FedSkillGit evolves one shared document under a
population-level acceptance test, and two extensions follow from that design. A
single document admits no per-client personalisation, so a layered artefact in
which a validated common core is complemented by per-client overlays is a natural
next step. The publication gate consumes validation behaviour, so a reward-free or
preference-driven gate would extend the protocol to tasks without an automatic
evaluator. Treating the revision as
the unit of collaboration may also allow the wider toolkit of version control,
including branching, review and provenance, to be applied to the parts of the
agent stack that are still edited by hand.

\bibliography{references}

\appendices
\section{Prompt Templates of the Protocol}
\label{app:prompts}

% The templates are set in a small typewriter face; long lines are broken so
% that no prompt line overflows the column.
\sloppy
\makeatletter
\renewcommand{\verbatim@font}{\scriptsize\ttfamily}
\makeatother
\fvset{breaklines=true,breakanywhere=true,breaksymbolleft={},breaksymbolright={}}

This appendix gives the prompt templates of \ourmethod. The executor prompt
drives task execution, the optimizer prompt produces the patch and the local
experience of a round, the merge prompt updates the global experience document
and aggregates the patches, and the two privacy prompts implement the sanitizer
and the audit. Commit packaging is derived by fixed rules rather than by a
prompt.

\subsection{Executor prompt}
\begin{promptbox}[title={Executor: task execution under the shared skill}]
\begin{verbatim}
You are solving one task instance from {dataset}.

The skill document below is your only guidance. Nothing
else in this conversation overrides it. Read it before
you begin, and consult it whenever you are unsure how
to proceed.

### Skill document
{skill_document}

### Task instance
{task_instance}

### Output requirements
- Reason before answering, then emit exactly one answer
  block.
- Use the answer format defined by the dataset:
  {answer_format}
- Do not refer to the skill document in your answer.
- You have at most {max_turns} turns and {max_tokens}
  output tokens.
\end{verbatim}
\end{promptbox}

\subsection{Optimizer prompt}
\begin{promptbox}[title={Optimizer: patch and local experience for one round}]
\begin{verbatim}
You are improving a shared skill document using the
evidence of one client. The client holds a private
training set; you see its execution trajectories, not
its raw data or its gold answers.

### Current skill document
{current_document}

### Execution trajectories on this client's slice
{trajectories}

### Global experience document
{global_experience}

### What to produce
1. Diagnosis. Identify why the executor failed on this
   slice. Ground every claim in the trajectories above.
2. Patch. Propose at most {edit_budget} edits over named
   rules. Each edit is one of:
   - append <rule>: add a new rule after the named rule
   - insert <rule>: insert a rule immediately after it
   - replace <rule>: replace the body of the named rule
   Every edit must cite the trajectory evidence that
   justifies it. Do not add a rule that the global
   experience document already records as rejected.
3. Local experience. Summarise this round as the failure
   patterns observed on this slice and the expected
   effect of the proposed edits.

Return one patch and one local experience. Do not edit
any rule that the patch does not name. Keep the total
document size under {size_ceiling} tokens.
\end{verbatim}
\end{promptbox}

\subsection{Merge prompt}
\begin{promptbox}[title={Merge: experience update and patch aggregation}]
\begin{verbatim}
You are the merge step of a federated skill evolution.
You receive the current skill document and the set of
commits submitted in this round. Each commit packages a
patch over named rules together with the local
experience of the client that produced it.

### Current skill document
{current_document}

### Commits from this round
{commits}

### Global experience document
{global_experience}

### Step 1: update the global experience document
Deduplicate the reported experience. Record what this
round established about the task, including any revision
that was proposed and found ineffective. The experience
document is context for the optimizer and for this step.
It is never written into the skill document.

### Step 2: aggregate the patches
Apply every patch to the current document and classify
each edit under one of four dispositions:
- keep: the edit is new and applies cleanly
- deduplicate: the edit repeats one already kept
- promote: the edit belongs in a shared rule
- reject: the edit is structurally invalid
An edit is dropped only when it duplicates a kept edit
or is structurally invalid. Do not rewrite client text
and do not average edits; the returned document must be
one that a participant proposed.

Return the updated experience document, the candidate
skill document, and one disposition record per submitted
edit.
\end{verbatim}
\end{promptbox}

\subsection{Sanitizer prompt}
\begin{promptbox}[title={Sanitizer: rewrite a commit before transmission}]
\begin{verbatim}
You are the privacy stage of a federated skill evolution.
Rewrite the payload below so that it carries no private
content.

### Payload
{commit_payload}

### What to remove
Entities, file paths, table and column identifiers,
dataset-specific literals, and any other value that
identifies the client or its data.

### How to rewrite
- Replace each removal with a typed placeholder, for
  example {ENTITY}, {PATH} or {IDENTIFIER}. Do not delete
  the surrounding sentence.
- Keep the structure of the payload: rule identifiers,
  edit operations and section headers must survive
  verbatim, so that the patch still applies.
- Do not summarise, explain or comment. Re-emit the
  payload field by field.
\end{verbatim}
\end{promptbox}

\subsection{Private-value extraction prompt}
\begin{promptbox}[title={Audit reference: private-value set for one client}]
\begin{verbatim}
You are given a sample of one client's private training
data. Extract the literal values whose appearance in text
sent to another party would constitute a leak.

### Data sample
{data_sample}

### Output
Return a flat list of literal strings, grouped by kind:
- path: file or directory paths
- identifier: table, column, or record identifiers
- other: any remaining client-specific literal

Copy the values exactly. Do not paraphrase, generalise or
complete them. Prefer a short and precise list over a
long one, because the list is used as a rejection test:
a transmitted payload that contains any listed value is
returned to the sanitizer.
\end{verbatim}
\end{promptbox}

\section{Limitations}
\label{app:limitations}

\ourmethod studies federated skill evolution as a lightweight alternative to pooling execution traces, and its scope is bounded in five ways.

\paragraph{The shared object is a single document.}
The protocol evolves one skill document that every client is scored against. This is what makes population-level validation possible, since the gate consumes one artefact that the whole population can certify. It also means that a client whose distribution differs sharply from the rest of the population can only be served by improving the common document rather than by receiving an artefact of its own. FederatedSkill takes the opposite trade, keeping a personalised library per client at the cost of an acceptance test that no population can jointly apply.

\paragraph{Validation behaviour.}
Publication is earned on the union of the clients' validation sets, so the protocol presumes that each client can score a candidate document on a held-out split. This is a mild requirement where the task has an automatic evaluator and a costly one where success is subjective or expensive to judge. In the latter case the gate would require a stronger scoring procedure, likely with a human in the loop.

\paragraph{Privacy is enforced on the payload, not against inference.}
The sanitizer and the audit constrain what a transmitted commit carries, and the audit consumes a frozen private-value set drawn from the client's own slice, so a verdict is reproducible. Neither stage bounds what an adversary can infer from the sequence of accepted revisions themselves. Where a formal privacy guarantee is required, the two stages should complement a mechanism such as differential privacy rather than replace it.

\paragraph{Commits are visible to the server.}
Each commit is sanitized before it leaves a client and the server merges sanitized commits. The protocol does not add secure aggregation or any other cryptographic hiding of the transmitted payload, so the server sees every participant's sanitized proposal. The privacy filter is therefore essential rather than optional.

\paragraph{Convergence cost grows with the population.}
The convergence round count rises with the number of clients and falls with the participation ratio, so a deployment with many clients faces a real cost in wall-clock rounds unless the participation ratio is raised. The protocol shares a single seed document across clients, so it also assumes the clients are working on the same problem rather than on unrelated ones.

\section{Experimental Protocol Details}
\label{app:protocol}

This appendix records the settings behind the results in Section~\ref{sec:exp} so that the runs can be reproduced from the paper.

\paragraph{Models.}
Three model families are evaluated, one per row group of Table~\ref{tab:main}, namely GPT-5.4-mini, Qwen3.5-flash and Gemini3-flash. Within each family the same model plays both roles, the executor and the optimizer, so a difference between two methods cannot be attributed to an asymmetric split between the two roles.

\paragraph{Datasets and splits.}
We use SearchQA for open-domain question answering, LiveMath for research-level mathematics multiple choice and DocVQA for document visual question answering, all as shipped and including each corpus's own seed skill document. Each dataset is partitioned so that no two clients hold the same examples, and each client's slice is drawn from a different part of the task distribution, producing a deliberately non-IID split across question type, difficulty band and answer format. The held-out evaluation set is disjoint from every client's slice.

\paragraph{Client and participation settings.}
The main comparison uses $10$ clients with a $30\,\%$ participation ratio, so three clients participate per round. The scheduler is fixed in advance rather than adapting to the observed scores, and a client that is not scheduled in a round contributes nothing to that round. All methods use the same per-round budget envelope and therefore expose the same amount of private data per round. The seed document is held fixed across every method.

\paragraph{Methods.}
All methods share the executor, the per-round exposure, the merge machinery wherever an aggregation step exists and the publication gate. They differ only in how the document is produced. \textit{No skill} queries the frozen executor with no document, which separates what the model already knows from what the evolved document adds. \textit{Independent} evolves one copy of the seed per client on that client's own data with no collaboration and reports the mean over clients. \textit{Centralized} trains one process on the pooled per-round exposure and measures candidates on $\vunion$. \textit{DocMerge} has each client upload exactly one mature document, which a strong optimizer merges in a single call, and the result is measured on $\vunion$. This baseline isolates iterative scheduling against a single-shot merge.

\paragraph{Metric.}
The reported quantity is the task success rate (TSR) on a sealed held-out split, graded per dataset with that dataset's own evaluator. Every paired comparison uses the same items within the same run, and the reported artefact is the document published at termination, which is a rule fixed before any run. The convergence round count is reported only in Figures~\ref{fig:rounds-clients} and \ref{fig:rounds-participation}, and is measured as the number of rounds up to and including the last accepted round before the terminating pair of rollbacks in Section~\ref{sec:validation}.

\typeout{get arXiv to do 4 passes: Label(s) may have changed. Rerun}

\end{document}